\documentclass[runningheads]{llncs}

\usepackage{eccv}

\usepackage{eccvabbrv}

\usepackage{graphicx}
\usepackage{booktabs}
\usepackage{multirow}
\usepackage{threeparttable}
\usepackage{rotating}
\usepackage{arydshln}
\usepackage{adjustbox}
\usepackage{pifont}

\usepackage[accsupp]{axessibility}  
\definecolor{lightbluee}{rgb}{0.0007, 0.44, 0.737}

\newcommand{\genq}{\textcolor{lightbluee}{\textit{GenQ }}}
\newcommand{\genqnoblank}{\textcolor{lightbluee}{\textit{GenQ}}}
\newcommand{\bftab}{\fontseries{b}\selectfont}
\newcommand{\fakeparagraph}[1]{\noindent{\bftab #1}}

\usepackage[pagebackref,breaklinks,colorlinks,citecolor=eccvblue]{hyperref}

\usepackage{orcidlink}

\def\equationautorefname~#1\null{Eq.~(#1)\null}
\newcommand{\aref}[1]{\hyperref[#1]{Appendix~\ref{#1}}} 

\usepackage{amsmath,amsfonts,bm}

\def\eqref#1{equation~\ref{#1}}

\def\1{\bm{1}}

\def\vo{{\bm{o}}}

\def\vs{{\bm{s}}}

\def\vu{{\bm{u}}}
\def\vv{{\bm{v}}}
\def\vw{{\bm{w}}}
\def\vx{{\bm{x}}}

\def\vz{{\bm{z}}}

\def\mI{{\bm{I}}}

\DeclareMathAlphabet{\mathsfit}{\encodingdefault}{\sfdefault}{m}{sl}
\SetMathAlphabet{\mathsfit}{bold}{\encodingdefault}{\sfdefault}{bx}{n}

\begin{document}
\title{\emph{GenQ}: Quantization in Low Data Regimes with Generative Synthetic Data} 

\titlerunning{\emph{GenQ}: Generative Synthetic Data for Quantization}

\author{Yuhang Li\inst{1}\orcidlink{0000-0002-6444-7253} \and
Youngeun Kim\inst{1}\orcidlink{0000-0002-3542-7720} \and
Donghyun Lee\inst{1} \and 
Souvik Kundu\inst{2} \and
Priyadarshini Panda\inst{1}\orcidlink{0000-0002-4167-6782}
}

\authorrunning{Y. Li, Y. Kim, D. Lee, S. Kundu, P. Panda.}

\institute{Department of Electrical Engineering, Yale University \and
Intel Labs, San Diego\\
\email{\{yuhang.li, youngeun.kim, donghyun.lee, priya.panda\}@yale.edu}\\
\email{souvikk.kundu@intel.com}
}

\maketitle

\begin{abstract}
In the realm of deep neural network deployment, low-bit quantization presents a promising avenue for enhancing computational efficiency. However, it often hinges on the availability of training data to mitigate quantization errors, a significant challenge when data availability is scarce or restricted due to privacy or copyright concerns. Addressing this, we introduce \genqnoblank, a novel approach employing an advanced Generative AI model to generate photorealistic, high-resolution synthetic data, overcoming the limitations of traditional methods that struggle to accurately mimic complex objects in extensive datasets like ImageNet.
Our methodology is underscored by two robust filtering mechanisms designed to ensure the synthetic data closely aligns with the intrinsic characteristics of the actual training data. In case of limited data availability, the actual data is used to guide the synthetic data generation process, enhancing fidelity through the inversion of learnable token embeddings.
Through rigorous experimentation, \genq establishes new benchmarks in data-free and data-scarce quantization, significantly outperforming existing methods in accuracy and efficiency, thereby setting a new standard for quantization in low data regimes. 
Code is released at \url{https://github.com/Intelligent-Computing-Lab-Yale/GenQ}.
\end{abstract}

\section{Introduction}
A variety of model compression techniques have been developed to deploy large deep learning models on embedded/mobile devices without significant accuracy drops. One prominent method is neural network quantization \cite{gholami2021survey}, which involves converting 32-bit floating-point models into a compact low-bit fixed-point format. This transformation leverages the efficiency of fixed-point computation and the benefits of reduced memory usage. Other notable techniques include pruning~\cite{blalock2020state, molchanov2016pruning} and knowledge distillation~\cite{hinton2015distilling}, both of which aim to reduce network size while preserving as much of the original model's performance as possible.

While these methods effectively accelerate neural networks, they typically require finetuning on the original training dataset to mitigate any accuracy loss in the compressed model. However, accessing the original training data can be challenging due to privacy concerns or Intellectual Property (IP) protection issues \cite{mehmood2016protection}. This challenge has led to the emergence of \emph{data-free quantization}~\cite{haroush2020knowledge, jung2019learning, cai2020zeroq}. This approach involves generating synthetic data from a pretrained network and then finetuning the compressed or quantized model using this synthetic data, thereby circumventing the need for original training data.

\begin{figure}[t]
    \centering
    \includegraphics[width=\linewidth]{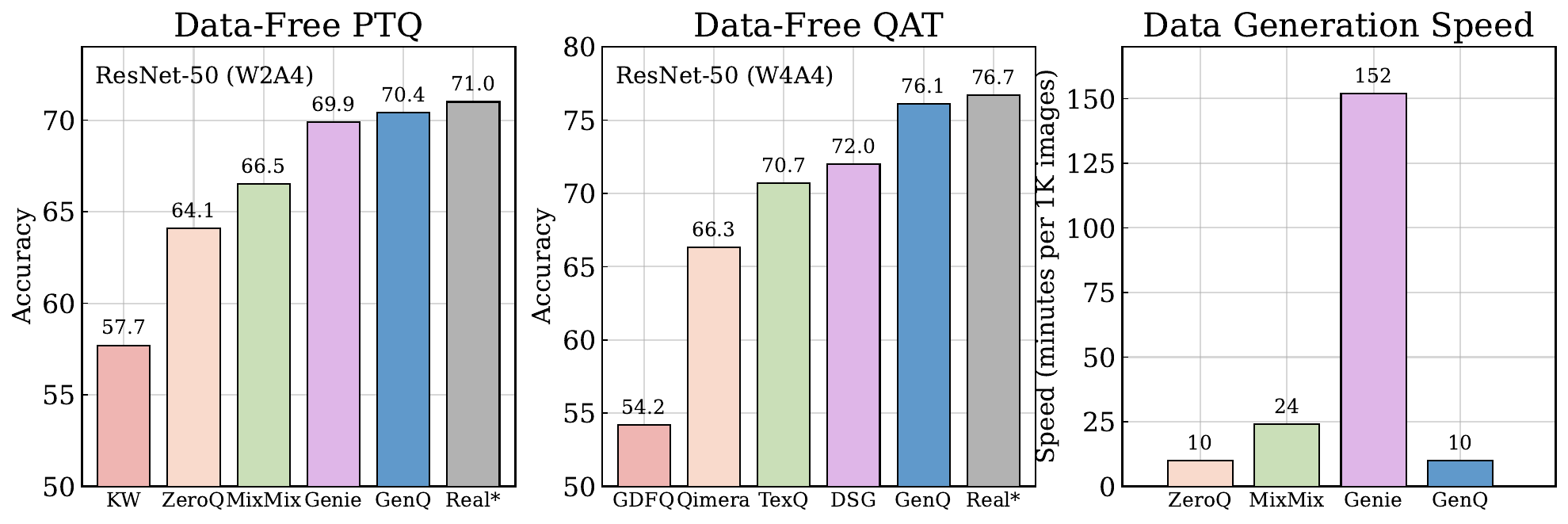}
    \caption{Comparison of \genq with existing methods on ImageNet. (1) Data-Free PTQ, (2) Data-Free QAT, (3) Data Generation Speed. Real* denotes using real ImageNet data in zero-shot quantization. }
    \vspace{-4mm}
    \label{fig_examples}
\end{figure}

Several prior works generate synthetic data by inverting the knowledge from pre-trained full-precision models. For instance, \cite{cai2020zeroq, haroush2020knowledge, zhang2021diversifying} propose to optimize the images by aligning them with the training data in terms of activation statistics. \cite{xu2020generative, liu2021zero} apply generative adversarial networks \cite{goodfellow2020generative} to synthesize the images. However, these approaches often struggle to match the compression performance achievable with real training data.
The primary challenge lies in the inherent complexity of reverse mapping from a lower to a higher dimension~\cite{li2021mixmix}. For example, in the context of the ImageNet dataset, this involves mapping from a 1000-dimensional space to an image space of $224\times 224\times 3$. Such transformation makes accurate recreation of complex objects a hard problem.
Additionally, these methods cost a significant amount of time to generate the images.

In this paper, we introduce \genqnoblank, a new pipeline utilizing the advanced Generative AI model such as Stable Diffusion~\cite{stablediffusion} to generate high-quality data for quantization purposes. This approach aligns with the growing trend of applying AI-generated content (AIGC) in deep learning and vision applications~\cite{zhang2023complete, bommasani2021opportunities, tian2023stablerep, wu2024datasetdm}. While direct application of generative AI data risks distribution shifts from original training data, we address this through two filtering mechanisms, ensuring the selection of in-distribution synthetic data.
Our method excels in data-free quantization scenarios.
More remarkably, in a majority of the model deployment scenarios with limited training data (\eg one image per class), \genq is significantly effective via prompt tuning to align generated images more closely with the real data. 
We conduct comprehensive experiments to validate the effectiveness of \genqnoblank, including both Post-Training Quantization (PTQ) and Quantization-Aware Training (QAT), on various neural network architectures (\eg CNNs~\cite{he2016deep, sandler2018mobilenetv2} and ViTs~\cite{dosovitskiy2020image}). \autoref{fig_examples} showcases  effectiveness of \genq on data-free quantization. Meanwhile, compared to other data generation approaches, our method achieves faster generation of synthetic data (up to 15$\times$ faster), increasing both efficiency and effectiveness. 

We summarize our contributions as follows:
\begin{enumerate}
    \item We propose \genqnoblank, the first work to leverage advanced text-to-image synthetic data for quantization in data-free scenarios. To reduce the distribution gap between synthetic data and real data, we introduce a suite of filtering mechanisms to select in-distribution synthetic data, including energy score filtering and BatchNorm Distribution filtering.
    \item In case of limited data availability, we also propose to learn the token embedding to guide the synthesis using real data. 
    \item We conduct extensive experiments to show the usefulness of \genq in generating synthetic data that helps achieve SoTA quantization performance in both PTQ and QAT. For example, our 4-bit QAT-based ResNet-50 achieves 76.10\% accuracy on the ImageNet dataset, outperforming the latest existing method~\cite{chen2023texq} by 5.4\%. 
\end{enumerate}

\section{Related Work}
\label{sec_related}
\fakeparagraph{Quantization Methods.}
Quantization approaches to compress pretrained Deep Neural Networks (DNNs) can be broadly divided into PTQ and QAT. 
PTQ performs \emph{calibration} on a pre-trained DNN after quantizing its weights and activations to low bits to maintain the original accuracy~\cite{nahshan2021loss, fang2020post, liu2021post}. For example, \cite{banner2019post, finkelstein2019fighting} propose bias correction in the convolutional layers after quantization. \cite{zhao2019improving} splits the outliers into additional channels to reduce the quantization error. Recently, a line of works~\cite{nagel2020up, li2021brecq, wei2022qdrop, hubara2020improving} leverage the weight rounding optimization to reconstruct the activation of the original model. 
While PTQ mostly adopts ad hoc strategies to prevent accuracy loss from quantization, QAT can significantly improve the accuracy of the quantized model through end-to-end finetuning. 
The success of QAT is mainly based on the usage of Straight-Through Estimator (STE)~\cite{yin2019understanding}, enabling gradient-based optimization~\cite{hubara2017quantized, choi2018pact, esser2019learned, zhou2016dorefa, zhang2018lq, yamamoto2021learnable}. 
QAT also explores optimizing the step size~\cite{esser2019learned}, clipping threshold~\cite{choi2018pact}, non-linear interval~\cite{jung2019learning}, non-uniform quantization levels~\cite{zhang2018lq} together with the weights. 

\fakeparagraph{Data in Quantization. }
To quantize a model to below 8 bits, data is essential, especially in QAT where end-to-end finetuning occurs. When access to real images is restricted due to privacy and copyright issues, ZeroQ~\cite{cai2020zeroq} proposes to synthesize images in replacement of real data. 
There are two categories of creating synthetic images: (1) Inverting the images directly via gradient descent~\cite{cai2020zeroq, zhang2021diversifying, li2021mixmix, haroush2020knowledge, jeon2023genie}, and (2) Learning a Generative Adversarial Network (GAN) to continuously generate images~\cite{xu2020generative, liu2021zero}. 

The image inversion~\cite{yin2020dreaming} relies on Batch Normalization (BN)~\cite{cai2020zeroq, haroush2020knowledge} as an optimization metric to distill the data and obtain high-fidelity images. 
Other inversion methods based on loss function~\cite{zhong2022intraq}, model ensembling~\cite{li2021mixmix}, and advanced convolutional layers~\cite{jeon2023genie} have been proposed to obtain better synthetic data. 
However, the inversion approach is limited in certain ways: (1) Inverting images adds a significant computation burden, which makes QAT with synthetic data prohibitive; (2) Models that do not have BN layers, such as ViTs, cannot invert images ~\cite{dosovitskiy2020image}.
The second category proposes to use GAN as an image synthesis engine~\cite{liu2021zero, xu2020generative, choi2021qimera, zhong2022intraq, qian2023adaptive}. These works finetune the generator and the quantized model simultaneously. 

Our method opens a third category for quantization in low data regime, \ie, using text-to-image diffusion models. Additionally, we investigate how to use limited real data to guide the generation of synthetic data. 
To the best of our knowledge, this scenario is being studied for the first time. 

\fakeparagraph{Synthetic Data in Deep Learning.}
Initial studies, such as those by \cite{besnier2020dataset, zhang2021datasetgan, jahanian2021generative}, explored the use of GANs for assisting DNNs in classifier tuning, object segmentation, and contrastive learning. Recently, the emergence of text-to-image models \cite{dhariwal2021diffusion, lugmayr2022repaint, rombach2022high, saharia2022palette} has revolutionized the synthesis of high-quality data in deep learning, owing to their effectiveness and efficiency.
For example, \cite{he2022synthetic} utilized GLIDE~\cite{nichol2021glide} for image synthesis in classifier tuning on the CLIP model~\cite{radford2021learning}. In a more advanced application, StableRep~\cite{tian2023stablerep} leverages Stable Diffusion to generate datasets for contrastive learning, using synthetic data from different seeds as positive pairs. Meanwhile, \cite{azizi2023synthetic} investigates the synthesis of data within the ImageNet label space observing accuracy improvements. Furthermore, Fill-up~\cite{shin2023fill} employs synthetic data from text-to-image models to balance the long-tail distribution in training datasets.

\section{Preliminaries}

\subsection{Quantization}
Uniform quantization maps full-precision weights into fixed-point numbers. For a step size $s\in\mathbb{R}$, the integer value of weights $\vw$ are given by
\begin{equation}
\vw _{\text{int}}= \mathrm{clip} \left (\left \lfloor \frac {\vw }{s}\right \rceil +\vz, n, p\right ),
\end{equation}
where $\lfloor \cdot \rceil$ denotes the rounding-to-nearest operation, and $n$ and $p$ represent the lower and upper bound of the integers, respectively. For example, under the $b$-bit uniform quantization, $n$ and $p$ are set to $0$ and $2^{b-1}$. The zero-point vector $z$ is an all-$z$ vector, where $z = -\lfloor \frac{\min(\vw)}{s}\rceil$. Thus, the quantized weights $\vw^q$ are given by
\begin{equation}
     \vw^q=s(\vw _{\text{int}}-\vz ).
     \label{eq_dequant}
\end{equation}

\fakeparagraph{Quantization-Aware Training (QAT). }
We utilize the Learned Step Size Quantization (LSQ)~\cite{esser2019learned} to update the step size $s$ and $\vw$ with gradient descent. The gradient to step size can be computed with Straight-Through Estimator (STE)~\cite{yin2019understanding}, \ie, $\frac{\partial \lfloor x\rceil}{\partial x} = 1$. QAT requires a significant amount of data and GPU resources to perform end-to-end fine-tuning. 

\fakeparagraph{Post-Training Quantization (PTQ). }
Under this PTQ regime, the training data is much less than QAT, hindering the learning of step size. To obtain the desired step size $s$ under PTQ, we perform layerwise quantization-error minimization~\cite{banner2019post}. 
In addition, we also conduct layer-wise tuning of the weight rounding during PTQ as did in~\cite{nagel2020up, li2021brecq, wei2022qdrop, jeon2023genie}, which can significantly improve the model performance with only 1K images.  

\subsection{Stable Diffusion Data Generation}
Stable Diffusion utilizes the Denoising Diffusion Probabilistic Model (DDPM)~\cite{ho2020denoising} to implement the training and inference. 
It contains a text encoder $\tau_\theta(\cdot)$, image encoder $\mathcal{E}(\cdot)$ and image decoder $\mathcal{D}(\cdot)$, and finally a denoising U-Net $\epsilon_\theta(\cdot)$.

\fakeparagraph{Inference Process. }
During inference, a random noise image latent is sampled from Gaussian distribution $\vz\sim\mathcal{N}(\mathbf{0, 1})$. Suppose the text prompt is $\vs$, the denoising U-Net fuses the text embedding $\tau_\theta(\vs)$ and the image and visual embedding through cross-attention layers and denoises the image latent gradually. 
After $T$ timesteps of denoising, the image latent is decoded by $\mathcal{D}(\cdot)$ to generate the high-resolution image. 

\fakeparagraph{Training process. }
Training the denoising U-Net essentially involves encoding the images first ($\hat{\vx}_0 = \mathcal{E}(\vx)$), and then diffusing the image through a Markov chain process $q(\hat\vx_t|\hat\vx_{t-1}) = \mathcal{N}(\hat\vx_t; \sqrt{1-\beta}\hat\vx_{t-1}, \beta I)$. Note that $t\in[1, T]$ is the number of diffusion timesteps and $\beta$ is a hyper-parameter to control the perturbation. When $t=T$, then $\hat{\vx}_t=\vz$. Now the denoising U-Net can be trained by a reverse process by matching the output and the noises added before:
\begin{equation}
    \min_{\theta} \mathbb{E}_{t\sim\mathcal{U}(1, T), \epsilon\sim\mathcal{N}(\bf{0}, \mI)} \lambda(t) ||\epsilon-\epsilon_\theta(\hat\vx_t, t)||_F^2,
\end{equation}
where $\lambda(t)$ is a positive weighting function~\cite{ho2020denoising}, $\epsilon$ is a noise vector predicted from $\vx_t$. 
We use Stable Diffusion v1-5~\cite{stablediffusion} in our pipeline. We refer to \cite{rombach2022high} for more details about its principle.

\section{Methodology}

In this section, we introduce our methodology to synthesize the training data with Stable Diffusion. 
We discuss two quantization scenarios in low data regimes: (1) {Data-Free Quantization} (DFQ) where only the label space and a pre-trained full-precision model are provided, and (2) Data-Scarce Quantization (DSQ) where limited training data is provided in addition to the pre-trained model and label space. 
We first discuss \genqnoblank's data generation method under DFQ and DSQ and then introduce the quantization techniques we used.

\begin{figure*}
    \centering
    \includegraphics[width=\textwidth]{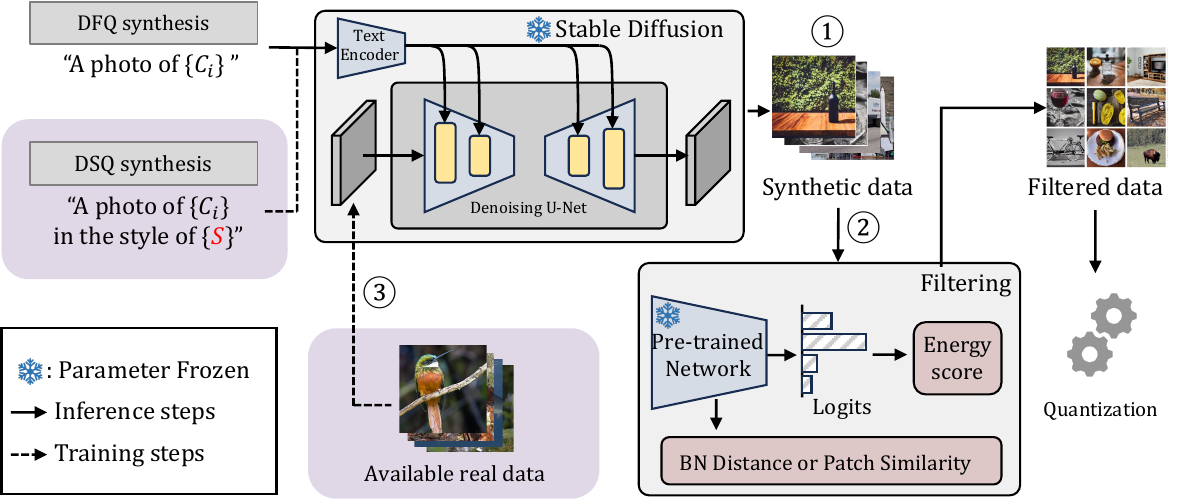}
    \caption{The overall image synthesis and filtering procedure of \genqnoblank. \raisebox{.5pt}
    {\textcircled{\raisebox{-.9pt} {1}}} For DFQ synthesis, we directly use the label as the text prompt, and 
    {\textcircled{\raisebox{-.9pt} {2}}} we use several metrics to filter our out-of-distribution synthetic images.  
\raisebox{.5pt}{\textcircled{\raisebox{-.9pt} {3}}} For data-scarce synthesis, real images are used to optimize prompt $\{\mathtt{S}\}$ (\autoref{eq:fewshot_opt}).  We then generate the synthetic images with the optimized prompt.}
    \label{fig_overview}
\end{figure*}

\subsection{Data-Free \textbf{\genq}}
We propose a \textit{label prompt} method that only uses the label as the prompt to synthesize the images without accessing the pre-trained model. 
For example, for an ImageNet pre-trained model~\cite{deng2009imagenet}, we can synthesize the images based on 1000 class names. 
\begin{equation}
    \mathrm{Prompt} = \texttt{"A photo of a \{D\} \{C\}"},
\end{equation}
where $\texttt{\{D\}}$ is a template adjective (\eg, ``nice, dark, small") derived from the CLIP ImageNet template~\cite{radford2021learning}\footnote{We list the full template in Appendix A}. $\texttt{\{C\}}$ is a randomly chosen class label name. As an example, suppose we choose the class \texttt{hamster}, then a random prompt can be generated as \texttt{"A photo of a small hamster"}. Subsequently, we directly use the text-to-image model like Stable Diffusion \cite{rombach2022high} to synthesize the data.  

In our \emph{label prompt} method, we generate the images relying solely on the label name of each class and the ability of Stable Diffusion (\autoref{fig_overview}{\textcircled{\raisebox{-.9pt} {1}}}). 
However, this operation does not leverage any prior knowledge in the original training data, which might cause a distribution shift from original training data to synthetic training data. 
To generate better-quality data for quantization, we propose \emph{model-dependent selection} to leverage the prior knowledge embedded in the pre-trained full-precision model. 

We propose a set of filtering mechanisms for model-dependent selection. We argue that the synthetic data selection can be treated as an Out-of-Distribution (OOD) detection problem~\cite{pimentel2014review}. However, contrary to OOD detection, we intend to select in-distribution data from the generated synthetic data for quantization purposes~(\autoref{fig_overview}{\textcircled{\raisebox{-.9pt} {2}}}).

\fakeparagraph{Energy Filtering. }
We use energy filtering to calculate the energy score~\cite{liu2020energy} as,
\begin{equation}
    E(\vx, f) = -\alpha \sum_{i=1}^C e^{-f_i(\vx)/\alpha},
\end{equation}
where $f_i(\vx)$ denotes the $i^{th}$ value of the network's output logits. $C$ and $\alpha$ denote the number of image classes and temperature, respectively. 
\cite{liu2020energy} has shown that the energy score is an OOD detector due to its theoretical connection with the likelihood function. The neural networks trained by cross-entropy loss inherently decrease the energy of the training data, hence, the OOD data will have relatively higher energy. 

To select synthetic images that have a similar quantization effect to that of real image data, we pass the synthetic data through the full-precision pre-trained model. Then, we calculate the energy scores of all synthetic images and only select those images for quantization that yield energy scores lower than a certain user-defined threshold. 
We will experiment with the choice of this threshold in \autoref{sec_ablation}.   

\fakeparagraph{BatchNorm Distribution Filtering. }
Originally proposed in \cite{yin2020dreaming, cai2020zeroq}, the channel mean and channel variance distance computed in the BN layers~\cite{ioffe2015batch} are regarded as a standard metric for optimizing the synthetic data in Convolutional Neural Networks (CNNs). 
We use this metric to evaluate the distance between synthetic images and original training images on the full-precision pre-trained model. 
We calculate the BN distance by
\begin{equation}
    D_{BN} = \sum_{\ell=1}^n \left(||\mu_\ell^s-\mu_\ell||_F + ||\sigma_\ell^s-\sigma_\ell||_F\right),
    \label{eq_bn_dist}
\end{equation}
where $\mu_l, \sigma_l$ denote the running mean and variance of the layer $l$ activation from original training images, and $\mu^s_l, \sigma^s_l$ are the current mean and variance of the layer $l$ activation from synthetic images.
Naively, for each generated synthetic data, we can evaluate the BN distance $D_{BN}$ and filter out the ones with large $D_{BN}$. 
However, we find this approach does not bring improvement in the final data quality due to the difference between single data and batched data. 
Evaluating $D_{BN}$ on a single image will lead to biased estimation as it ignores its interaction with other data. 
Ioffe \cite{ioffe2017batch} also demonstrates that with small batch sizes, the estimation of the batch mean and variance used during training become a less accurate approximation of the mean and variance used for testing.

To deal with this problem, we define a \emph{BN sensitivity} metric, which measures the independence of one image from other images in a batch. Formally, given a batch of input images $\{\vx_i\}_{i=1}^B$ where $B$ indicates the batch size, we define the BN sensitivity of the $i$-th image $S(\vx_i)$ as
\begin{equation}
    S(\vx_i) = D_{BN}(\{\vx_j\}_{j=1}^B) - D_{BN}(\{\vx_j\}_{j=1, j\neq i}^B).
\end{equation}
Here, the BN sensitivity indicates the change in BN distance after removing the selected image. As such, a large BN sensitivity means the current image could potentially damage the internal distribution when it is batched with other images.

\fakeparagraph{Patch Similarity Filtering for ViTs. }
As we discussed in \autoref{sec_related}, since ViTs do not have BN layers, we adopt the patch similarity metric in \cite{li2022patch} to do additional filtering of the synthetic images after the energy filtering. Formally, given the output features tensor $\vo$, we first calculate the cosine similarity matrices between any two sub-tensors in the patch dimension, given by
\begin{equation}
    \Gamma(\vo_i, \vo_j) = \frac{\vo_i \cdot \vo_j}{||\vo_i|| \ ||\vo_j||},
\end{equation}
where $\vo_i$ denotes the $i$-th patch of the output featuremap. 
After calculation, we get the $N\times N$ ($N$ is the \#patches) matrix $\mathbf{\Gamma}$. To measure the diversity of the patch similarities, we calculate differential entropy as follows
\begin{equation}
    H = -\int \hat{f}_h(x) \cdot \log [\hat{f}_h(x)] dx. 
\end{equation}
where $\hat{f}_h(x)$ is the continuous probability density function of $\mathbf{\Gamma}$, which can be obtained by the kernel density estimation~\cite{li2022patch}. 
A low entropy value indicates that the patch similarity distribution is less diverse. Hence, we can select the synthetic data that has the highest diversity in patch similarity. 

\fakeparagraph{Filtering Pipeline. }
The overall filtering pipeline is a two-stage procedure. First, regardless of the type of pre-trained model, we apply energy score filtering. Second, based on the category of the model (CNNs or ViTs), we either apply BatchNorm Distribution Filtering or Patch Similarity Filtering. 
 The selected synthetic data is then used for quantization. 

\subsection{Data-Scarce \textbf{\genq}}
\label{sec_dsq}
In this section, we describe how to synthesize and select synthetic data given a limited amount of training data (\ie, data-scarce quantization). This case could be more common than the DFQ scenario as in practice it is not hard to obtain some permitted training data. {We hypothesize that synthesizing a number of images based on this limited dataset can enhance quantization performance.}

Considering the ImageNet-1k dataset~\cite{deng2009imagenet} as the original training dataset, we assume that one image per class (\ie, 1-shot) can be accessed by the cloud server. 
In this case, we propose to synthesize the images utilizing the existing information from the training data. 
Specifically, we optimize a text token embedding $\{\mathtt{S}\}$ with the prompt shown below 
\begin{equation}
        \mathrm{Prompt}_{[i]} = \texttt{"A photo of a }\{\mathtt{C}_{[i]}\}\ 
        \texttt{in the style of \{}\mathtt{S}\texttt{\}"}.
\end{equation}
Here, the learnable token embedding $\{\mathtt{S}\}$ indicates the dataset characteristics of ImageNet. 
To optimize this learnable token embedding, we associate each class name and the corresponding image into pairs ($\{\mathtt{C}_{[i]}, \vx_{[i]}\}$, where $[i]$ is the class index), and let the Stable Diffusion generate $\vx_{[i]}$ given $\mathrm{Prompt}_{[i]}$. The optimization objective is given by:
\begin{equation}
    \mathbb{E}_{t\sim\mathcal{U}(1, T), i\sim\mathcal{U}(1, M), \epsilon\sim\mathcal{N}(\bf{0}, \mI)} \lambda(t) ||\epsilon-\epsilon_\theta(\vx_{[i]t}, t, \mathrm{Prompt}_{[i]})||_F^2,
    \label{eq:fewshot_opt}
\end{equation}
where $M$ is the number of all object classes. 
This method learns the token across multiple objects, aiming to characterize the whole dataset\footnote{We provide more technical details in Appendix A. }. 
\autoref{fig_overview}{\textcircled{\raisebox{-.9pt} {3}}} illustrates the data-scarce prompt optimization. After optimization, we directly used the learned token embedding and the class label to generate images and filter the output images using the same technique in \autoref{fig_overview}{\textcircled{\raisebox{-.9pt} {2}}}.

\subsection{Quantization with Synthetic Data from \genq}

The selected synthetic data from \genq is then used to quantize a full-precision pre-trained model using QAT or PTQ. For PTQ, we adopt the state-of-the-art reconstruction-based rounding optimization methods~\cite{li2021brecq, jeon2023genie}. As for QAT, we propose to finetune the quantized network on top of a PTQ model. 
Specifically, after the rounding optimization in PTQ, the quantization becomes
\begin{equation}
\vw _{\text{int}}= \mathrm{clip} \left (\left \lfloor \frac {\vw }{s}\right \rfloor + \mathrm{sgn}(\vv) +\vz, n, p\right ),
\end{equation}
where $\mathrm{sgn}(\vv)$ is the learned rounding indicating up or down, which is already well-optimized in the PTQ reconstruction stage. 
To initialize from the PTQ model and stabilize the training process, we freeze all previous learnable variables including $\vw$ and $\vv$, and reinitialize an all-zero vector $\vu$. The new quantization function is thus given by
\begin{equation}
    \vw _{\text{int}}= \mathrm{clip} \left (\left \lfloor \frac {\vw }{s}\right \rfloor + \mathrm{sgn}(\vv) + \left \lfloor \frac {\vu }{s}\right \rceil +\vz, n, p\right ).
\end{equation}
The de-quantization step remains the same with \autoref{eq_dequant}. 
By introducing an additional variable $\vu$ and freezing the previous variable, we start from the original PTQ model and avoid passing the gradient through floor operation. As a result, the $\vu$ can be safely updated through STE in finetuning. Note, during PTQ or QAT, we only use the synthetic data from \genq to perform quantization.

\section{Experiments}

In this section, we empirically demonstrate the effectiveness and the efficiency of our \genq synthetic data in both a qualitative and a quantitative way. 
For the PTQ scenario, we follow the conventional setup~\cite{li2021brecq, jeon2023genie, li2021mixmix} to synthesize 1k images. As for the QAT scenarios, we synthesize 1.2M images with 1200 images in each object class to match the original ImageNet dataset volume. 
Unless specified, we use Stable Diffusion v1-5~\cite{huggingfaceRunwaymlstablediffusionv15Hugging} and set the guidance scale to 3.5. We will first provide the visualization of our synthetic data, the latency comparison of generating the data, and then compare them against existing state-of-the-art methods across multiple setups. 
Finally, we analyze \genq by conducting various ablation studies. All experiments and accuracy noted are for the ImageNet dataset.  

\begin{figure}[t]
    \centering
    \includegraphics[width=\linewidth]{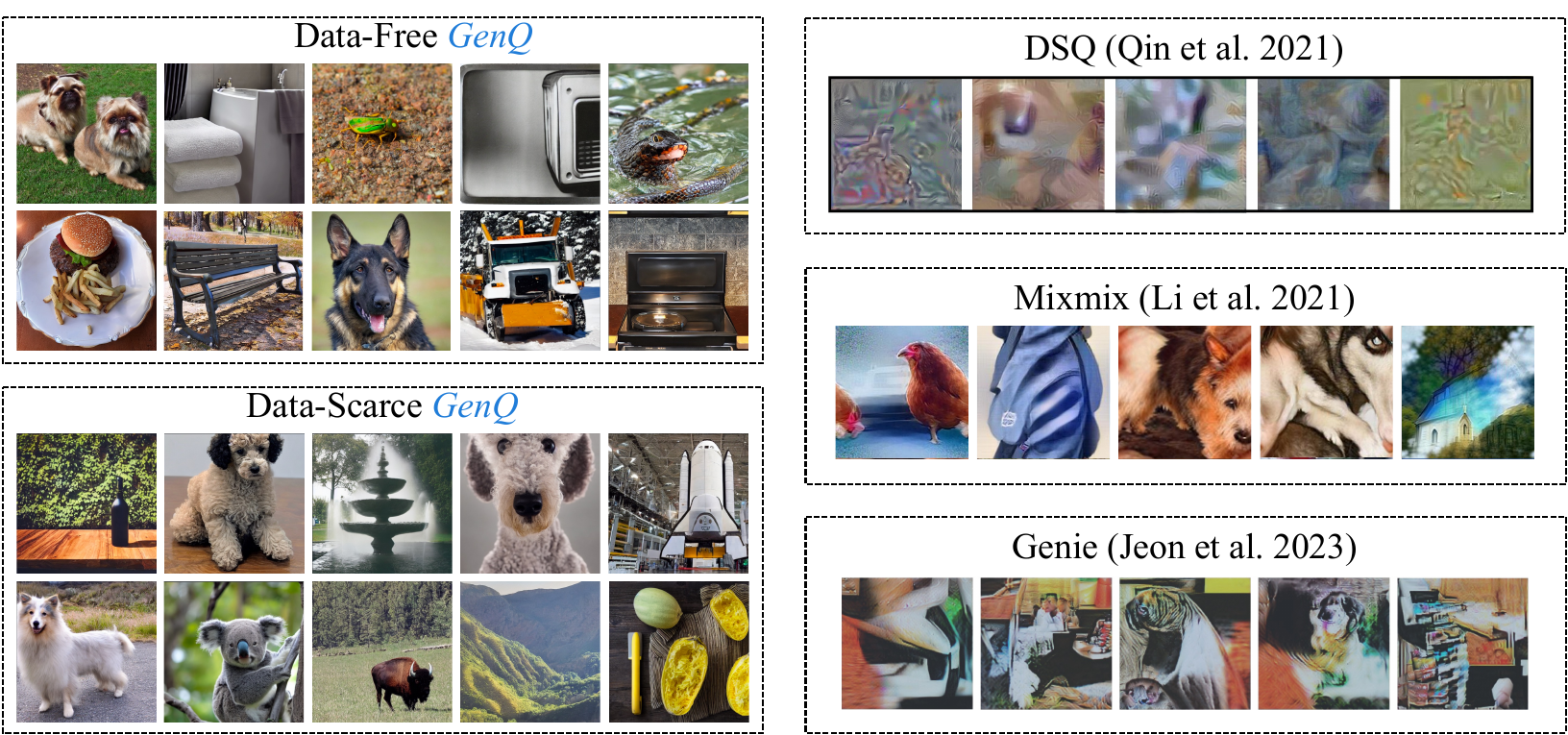}
    \caption{Visualization of DF/DS \genq and existing data synthesis method for quantization in low data regimes. }
    \label{fig_compare}
\end{figure}

\subsection{Analysis on Synthetic Data }

\fakeparagraph{Visualization. }
As shown in \autoref{fig_compare}, the \genq images after filtering are high-quality and visually similar to real images. 
To compare the synthetic data with existing data-free quantization, we also provide the images from \cite{qin2023diverse, li2021mixmix, jeon2023genie}.  
It can be observed that our method can generate much higher quality synthetic data than previous inversion-based methods. For example, \cite{qin2023diverse} only demonstrates certain textures rather than objects, \cite{li2021mixmix} shows some objects but the scene shows less resolution and clarity. 

\fakeparagraph{Data Synthesis Cost. }
We further study the cost of each synthetic data generation method. Specifically, we measure the latency of Genie~\cite{jeon2023genie}, TexQ~\cite{chen2023texq}, and our method corresponding to image inversion, GAN-based, and text-to-image synthesis approaches, respectively, for 1000 images on the ImageNet dataset. 
The pre-trained model used for filtering or input optimization is ResNet-50. 
We report the latency in \autoref{tab_latency}, from which we can find using text-to-image synthesis provides faster data generation than image inversion-based techniques. 
Although GAN-based data generation is faster than \genq, their data is not reusable across model types, which increases the overall latency if many models have to be quantized.

\begin{table}[t]
\caption{Comparison of synthetic data generation approaches w.r.t. (a) generation speed and (b) data quality.}
\vspace{-2em}
\begin{subtable}[t]{0.48\linewidth}
\scriptsize
\caption{\footnotesize Latency}
\centering
\begin{tabular}[t]{l|ccc}
Method & Type & Latency & Reusable \\
\hline
\textsc{Genie}~\cite{jeon2023genie} & Image inversion & 152 min & \ding{51} \\
TexQ~\cite{chen2023texq} & GAN-based &  10 min & \ding{55} \\
\genq & Text-to-image & 24 min & \ding{51}\\
\end{tabular}
\label{tab_latency}
\end{subtable}
\hspace{\fill}
\begin{subtable}[t]{0.48\linewidth}
\scriptsize
\caption{\footnotesize BatchNorm Distance}
\centering
\begin{tabular}[t]{l|cc}
Data Source & Res50 & MBV2 \\
\hline
\textsc{Genie}~\cite{jeon2023genie} & 0.014$\pm$0.001 & 1.03$\pm$0.296\\
TexQ~\cite{chen2023texq} & 0.045$\pm$0.013 & N/A \\
\genq & 0.024$\pm$0.002 & 0.78$\pm$0.106 \\
\end{tabular}
\label{tab_quality}
\end{subtable}
\vspace{-1em}
\end{table}

\fakeparagraph{Data Quality Assessment. }
We further measure the quality of the synthetic data through some quantitative evaluation metrics. Specifically, we measure the average BN distances of the generated data as well as the original training data on ResNet-50 and MobileNetV2. The results are shown in \autoref{tab_quality}, from which we can find that \genq largely closes the gap between synthetic data and real training data.

\newcommand{\genie}{\textsc{Genie}}

\begin{table}[tb]
\renewcommand{\arraystretch}{1.2}
\scriptsize
\centering
\caption{Evaluation of data-free PTQ on CNN models (top-1 accuracy (\%)).}
\begin{threeparttable}\vspace{-0.25cm}
\begin{tabular}{llclllll}
\hline
Quant. Method \:\:\:& Syn. Method \:\:\:& \#Bits W/A \:\:\:& \multicolumn{1}{c}{Res18} \:\:\:& \multicolumn{1}{c}{Res50} \:\:\:& \multicolumn{1}{c}{MBV2} \:\:\:& \multicolumn{1}{c}{MB-b} \:\:\:& \multicolumn{1}{c}{MNas-1}  \\ \hline
{Full Prec.} & N/A  & 32/32   & {71.08}    & {77.00} & {72.49} & {74.53} & {73.52}\\ \hline
\multirow{8}{*}{\textsc{Brecq} \cite{li2021brecq}} & Real Data  & \multirow{8}{*}{4/4}  & 69.62 & 75.45 & 68.84 & - & -\\
& ZeroQ$^\ddagger$~\cite{cai2020zeroq}  &  & 69.32 & 73.73 & 49.83 & 55.93 & 52.04\\
& KW$^\ddagger$~\cite{haroush2020knowledge}  &   & 69.08 & 74.05 & 59.81 & 61.94 & 55.48\\
& IntraQ~\cite{zhong2022intraq} &   & 68.77 & 68.16 & 63.78  & - & - \\
& Qimera~\cite{choi2021qimera} &   & 67.86 & 72.90 & 58.33  & - & - \\
& MixMix$^\ddagger$~\cite{li2021mixmix}    &   & 69.46 & 74.58 & 64.01 & 65.38 & 57.87  \\
& \genie{-D} & &  69.40 & 75.35 & 67.81 & 64.24 & 65.02\\
& \genq &  & \textbf{69.52} & \textbf{75.47} & \textbf{68.34} & \textbf{67.08} & \textbf{66.33} \\
\cdashline{1-8}
 & Real Data &    & {69.82}  & {75.51} & {69.11} & 69.26 & 68.40\\
{\genie{-M}$^*$} \cite{jeon2023genie} & {\genie-D} \cite{jeon2023genie} & 4/4  & {69.72} & \textbf{75.61} & 68.62 & 67.31 & 67.03 \\
& \genq   & &\textbf{69.77} & 75.50 & \textbf{68.96} & \textbf{68.74} & \textbf{68.06} \\
\hline
\multirow{7}{*}{\textsc{Brecq}~\cite{li2021brecq}} & Real Data& \multirow{7}{*}{2/4} & 65.25 & 70.65 & 54.22 & - & -\\
&   ZeroQ~\cite{cai2020zeroq} &  & 61.63 & 64.16$^\ddagger$ & 34.39 & 23.53 & 13.83 \\
  & KW$^\ddagger$~\cite{haroush2020knowledge}   &  & - & 57.74 & - & - & -\\
 & IntraQ~\cite{zhong2022intraq} &   & 55.39 & 44.78 & 35.38  & - & - \\
   & Qimera~\cite{choi2021qimera} &   & 47.80 & 49.13 & 3.73  & - & - \\
   &MixMix$^\ddagger$~\cite{li2021mixmix}  &   & - & 66.49 & - & - & -  \\
   & {\genie-D}~\cite{jeon2023genie} &   & 63.93 & 69.72 & 49.75  & 38.01 & 45.53 \\
  &  \genq  & & \textbf{65.04} & \textbf{69.90} & \textbf{53.08} & \textbf{47.31} & \textbf{50.84} \\
\cdashline{1-8}
& Real Data  &  & 66.05 & 70.96 & 56.42 &  55.00 & 54.66\\
\genie{-M}$^*$ \cite{jeon2023genie} & \genie{-D}~\cite{jeon2023genie}  & 2/4  & {64.86} & {69.89} &{51.47} & 47.69 & 48.38 \\
& \genq &  &  \textbf{65.72} & \textbf{70.35} & \textbf{54.82} & \textbf{52.77} & \textbf{52.76}\\
\hline
\end{tabular}
\begin{tablenotes}
\item[$\ddagger$] The figures are taken from ~\cite{li2021mixmix}.
\item[$*$] Denotes our implementation based on the open-source code.
\end{tablenotes}
\end{threeparttable}
\label{tab:ptq}
\end{table}

\begin{table}[tb]
\renewcommand{\arraystretch}{1.2}
\scriptsize
\centering
\caption{Evaluation of data-free QAT on CNN models (top-1 accuracy (\%)).}
\begin{threeparttable}\vspace{-0.25cm}
\begin{tabular}{lcccllll}
\hline
Method \:\:\:& \#Real Data \:\:\:&  \#Syn Data  \:\:\:&  \#Bits W/A \:\:\:&  Res18 \:\:\:&  MBV2 \:\:\:&  Res50  \\ \hline
LSQ~\cite{esser2019learned} & 1.2M & 0 & 4/4 & 71.10 & 69.50  & 76.70 \\
\cdashline{1-7}
GDFQ~\cite{xu2020generative} & 0 & 1.2M & \multirow{8}{*}{4/4} & 60.60 & 59.43 & 54.16 \\
ZAQ~\cite{liu2021zero} & 0 & 4.6M & & 52.64 & 0.10 & 53.02 \\
Qimera~\cite{choi2021qimera}  & 0 & 1.2M & & 63.84 & 61.62 & 66.25 \\
IntraQ~\cite{zhong2022intraq}  & 0 & 5k & & 66.47 & 65.10 & - \\
ARC+AIT~\cite{choi2022s}  & 0 & 1.2M & & 65.73 & 66.47 & 68.27 \\
DSG~\cite{qin2023diverse} & 0 & 1.2M & &  62.18 & 60.46 & 71.96 & \\
AdaDFQ~\cite{qian2023adaptive} & 0 & 1.2M & & 66.53 & 65.41 & 68.38 \\
TexQ~\cite{chen2023texq}  & 0 & 1.2M & &67.73 & 67.07 & 70.72\\
\genq & 0& 1.2M & & \bf 70.03 & \bf 69.65 & \bf 76.10 \\
\hline
LSQ~\cite{esser2019learned}  & 1.2M & 0 & 3/3 & 70.20 & 65.30 & 75.80 \\
\cdashline{1-7}
GDFQ~\cite{xu2020generative}  & 0 & 1.2M & \multirow{4}{*}{3/3} & 20.23 & 1.46 & 0.31 \\
AdaDFQ~\cite{qian2023adaptive} & 0 & 1.2M& & 38.10 & 28.99 & 17.63 \\
TexQ~\cite{chen2023texq} & 0 & 1.2M & &50.28 & 32.80 & 25.27\\
\genq & 0 & 1.2M & & \bf 68.18 & \bf 59.15 & \bf  73.99 \\
\hline
\end{tabular}
\end{threeparttable}
\label{tab_qat}
\end{table}

\subsection{Quantization Performance Evaluation}
In this section, we test the quantization accuracy with our synthetic data and compare it with various existing methods. We will introduce the experiment setup in each scenario individually.

\fakeparagraph{Comparisons on Data-Free PTQ. }
We start with evaluating our proposed method by testing it on data-free PTQ including (1) CNNs such as ResNet~\cite{goodfellow2016deep}, MobileNet~\cite{howard2017mobilenets, sandler2018mobilenetv2}, and MnasNet~\cite{tan2018mnasnet} and (2) Vision Transformers like original ViT~\cite{dosovitskiy2020image}.
For PTQ on CNNs, we employ the state-of-the-art PTQ algorithms, \textsc{Brecq}~\cite{li2021brecq} and \genie{-M}~\cite{jeon2023genie} to perform the W4A4 and W2A4 quantization, which reconstruct the activation output in a block-wise manner. In this case, we use 1024 synthetic images for quantization evaluation. 
For PTQ on ViT, we perform PTQ4ViT~\cite{yuan2022ptq4vit} and RepQ-ViT~\cite{li2023repq} across 5 different models, including ViT-M~\cite{dosovitskiy2020image}, DeiT-M~\cite{touvron2021training}, and Swin-B~\cite{liu2021swin}. We follow the open-source code implementation and generate 32 images as the calibration dataset to obtain the quantized ViTs. 
For all cases, we generate 2$\times$  synthetic data and filter 50\% of them to match the volume.

For CNNs, we select SOTA existing methods including ZeroQ~\cite{cai2020zeroq}, the Knowledge Within~\cite{haroush2020knowledge}, IntraQ~\cite{zhong2022intraq}, Qimera~\cite{choi2021qimera}, MixMix~\cite{li2021mixmix}, \genie{-D}~\cite{jeon2023genie}. 
For reference, we also include the performance of PTQ using 1024 real ImageNet-1k images.
We summarize all the accuracy results in \autoref{tab:ptq}. 
It can be observed that our \genq achieves the highest accuracy in most cases. Interestingly, we find that the network architecture affects the performance of synthetic data. On ResNet architectures, \genq and \genie{-D} have similar accuracy (0.1-0.5\%) levels, however, \genq largely outperforms other data-free algorithms on lightweight network architectures due to their lower quantization resilience. 
As an example, \genq increases 5\% accuracy compared to \genie{-D} on MobileNet-b W2A4 quantization and 4.4\% accuracy on MnasNet W2A4 quantization.

\begin{figure}[t]
\noindent\begin{minipage}{\textwidth}
\begin{minipage}{0.32\textwidth}
\centering
\includegraphics[width=\linewidth]{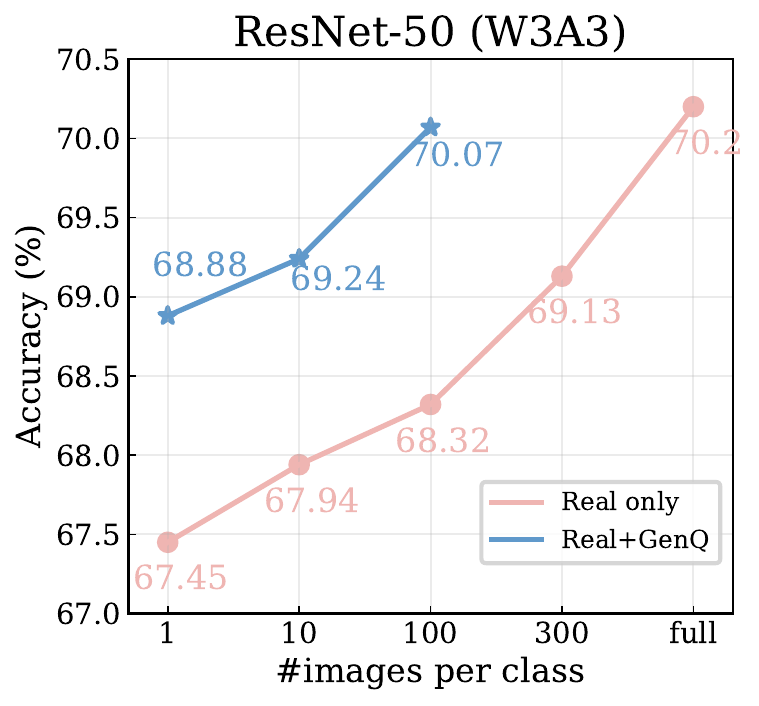}
\vspace{-7mm}
\captionof{figure}{Evaluation of data-scarce \genq.}
\label{fig_dsq}
\end{minipage}
\hfill
\begin{minipage}{0.65\textwidth}
\centering
\captionof{table}{Evaluation of data-free PTQ on ViTs (top-1 accuracy (\%)) }
\label{tab_ptq_vit}
\begin{adjustbox}{max width=\linewidth}
\begin{threeparttable}
\begin{tabular}{llcrrr}
\hline
Quant. Method & Syn. Method & \#Bits W/A  & \multicolumn{1}{c}{ViT-B} &\multicolumn{1}{c}{DeiT-B} & \multicolumn{1}{c}{Swin-B}  \\ \hline
{Full Prec.} & N/A  & 32/32 & 84.54 & 81.80 & 85.27\\ \hline
\multirow{3}{*}{PTQ4ViT~\cite{yuan2022ptq4vit}} & Real Data  & \multirow{3}{*}{4/4}  & 58.07 & 63.57 & 75.20\\
& PSAQ-ViT~\cite{li2022patch} & &  60.26 &  65.18 & 72.12 \\
& \genq &  & \textbf{63.17} & \textbf{67.08} & \textbf{74.27} \\
\cdashline{1-6}
\multirow{3}{*}{RepQ-ViT~\cite{li2023repq}} & Real Data  & \multirow{3}{*}{4/4}  &  67.93 &75.99 & 72.80 \\
& PSAQ-ViT~\cite{li2022patch} & &  60.18 &  74.69 & 54.21 \\
& \genq &  &  \textbf{67.50} & \textbf{76.10} & \textbf{70.08} \\
\hline
\end{tabular}
\end{threeparttable}
\end{adjustbox}
\end{minipage}
\end{minipage}
\vspace{-1em}
\end{figure}

Given that there is only one data-free algorithm for ViTs, PSAQ-ViT~\cite{li2022patch}, we compare \genq with PSAQ-ViT in Table \ref{tab_ptq_vit}. \genq consistently outperforms PSAQ-ViT~\cite{li2022patch}. 
For example, on ViT-B and Swin-B, our method has 7.3\% and 15\% accuracy improvement using the RepQ-ViT method. This result proves that \genq can achieve state-of-the-art performance on both CNNs and ViTs. 

\fakeparagraph{Comparisons on Data-Free QAT. }
We then compare \genq with existing data-free QAT methods, such as GDFQ~\cite{xu2020generative}, ZAQ~\cite{liu2021zero}, Qimera~\cite{choi2021qimera}, IntraQ~\cite{zhong2022intraq}, ARC~\cite{choi2022s}, AdaDFQ~\cite{qian2023adaptive}, TexQ~\cite{chen2023texq}.
Note that these methods jointly optimize GAN and quantized model. Thus, they can generate \emph{unlimited} synthetic data during finetuning. 
We originally have 1.6M synthetic images (1600 images/class), and then filter 0.4M (400 images/class) images using energy score and BN distance filtering. We use the SGD optimizer with a learning rate of 0.001 followed by a cosine annealing decay schedule for 50 epochs of QAT. 
Additionally, we report the LSQ results using 1.2M real ImageNet data. 
The results are summarized in \autoref{tab_qat}. 
Our method largely outperforms other data-free QAT methods and nearly approaches the accuracy of LSQ baseline. 
Remarkably, \genq improves the accuracy of W3A3 quantization by a large margin, for instance, \genq exceeds the accuracy of TexQ, the best-performing method, by 48\% on ResNet-50.

\fakeparagraph{Comparisons on Data-Scarce QAT. }
In this section, we evaluate the data-scarce QAT scenarios. Our main comparison is LSQ~\cite{esser2019learned} using different amounts of real data. We show how much of the real data can our \genq generated synthetic data match in practice. We experiment this on a ResNet-50 with W3A3 quantization. We initialize the QAT model by performing \genie{-M} on 1k real data and then finetune the model using \{1k, 10k, 100k, 300k, 1.2M\} real data, corresponding to 1-shot, 10-shot, 100-shot, 300-shot, and full-shot data regimes, respectively.
To generate \genq data, we initialize the token embedding as \texttt{ImageNet}, and train the embedding for 50k iterations, (see Appendix A for more details). 

As demonstrated in \autoref{fig_dsq}, the performance of QAT is highly correlated with the number of real data. Our method can boost the performance of QAT under the low data regime. For example, when only 1k real images are provided (\ie, 1-shot), \genq can achieve 68.88\% ImageNet accuracy, similar to the QAT performance that needs 300$\times$ more training data.

\subsection{Data Transferability Evaluation}

\begin{figure}[t]
\noindent\begin{minipage}{\textwidth}
\begin{minipage}{0.32\textwidth}
\centering
\includegraphics[width=\linewidth]{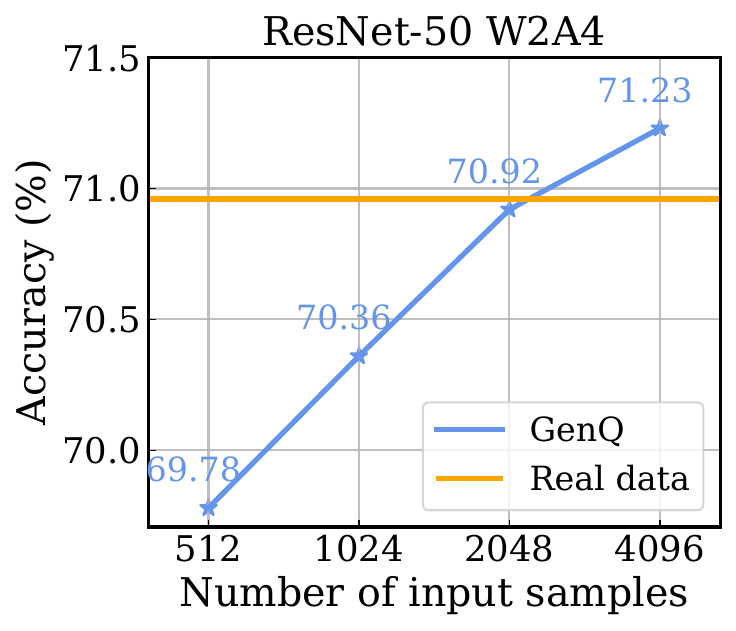}
\vspace{-7mm}
\captionof{figure}{Accuracy vs \# syn. data with data-free PTQ using Genie-M.}
\label{fig_data_num}
\end{minipage}
\hfill
\begin{minipage}{0.65\textwidth}
\centering
\captionof{table}{Cross-model evaluation on synthetic data. We compare the data transferability between (a) existing methods and (b) \genq. }
\begin{subtable}[t]{0.48\linewidth}
\footnotesize
\caption{\footnotesize \genie{-D} \& PSAQ-ViT}
\begin{tabular}[t]{l|cc}
Source model & Res18 & MBV2 \\
\hline
Res18 & 64.86 & 48.37\\
MBV2 & 64.21 &  51.47\\
ViT-B-16 & 47.27 & 17.96 \\
\end{tabular}
\label{tab_transfer1}
\end{subtable}
\hspace{\fill}
\begin{subtable}[t]{0.48\linewidth}
\footnotesize
\caption{\footnotesize \genq}
\begin{tabular}[t]{l|cc}
Source model & Res18 & MBV2 \\
\hline
Res18 & 65.72 & 54.32 \\
MBV2 & 65.52 & 54.82\\
ViT-B-16 & 65.44 & 54.24 \\
\end{tabular}
\label{tab_transfer2}
\end{subtable}
\end{minipage}
\end{minipage}
\vspace{-2em}
\end{figure}

It is demonstrated that the synthetic data extracted from one model has low transferability on other models~\cite{li2021mixmix}. 
To demonstrate that \genq has relatively high transferability, we conduct a cross-model evaluation, \eg, synthesizing images based on model A, and evaluate the quantization on model B. 
In Table \ref{tab_transfer1} we provide the results of \genie{-D} and PSAQ-ViT, tested on data-free PTQ W2A4 scenarios. We then summarize the results of \genq in Table \ref{tab_transfer2}. We use Genie-M as our PTQ method for CNN architectures and PASQ-ViT for ViT.
We observe that the \genq data, although filtered by a different model, only drops 0.2-0.5\% final accuracy, while the existing methods drop 0.6-33\% accuracy, especially when using ViT synthetic data for MobileNetV2 (MBV2) quantization. 

\subsection{Ablation study}

\label{sec_ablation}
In this section, we conduct ablation studies on two variables, (1) the number of synthetic data, and (2) the image filtering strength. 

\fakeparagraph{Number of synthetic data in PTQ. }
In practice, synthesizing images using \genq requires very low latency 
(1 image/second). Hence, we can safely increase the number of synthetic data with minimal overhead in PTQ cases. Yet retrieving more real data seems much more difficult if the training data is private or under IP protection. 
In \autoref{fig_data_num}, we show that increasing the synthetic data can consistently improve the PTQ performances of W2A4 quantized ResNet-50. \genq even outperforms the quantized model baseline (optimized using PTQ on the 1k real training data) when using only 4k synthetic input images.

\begin{figure}[t]
    \centering
    \includegraphics[width=0.8\linewidth]{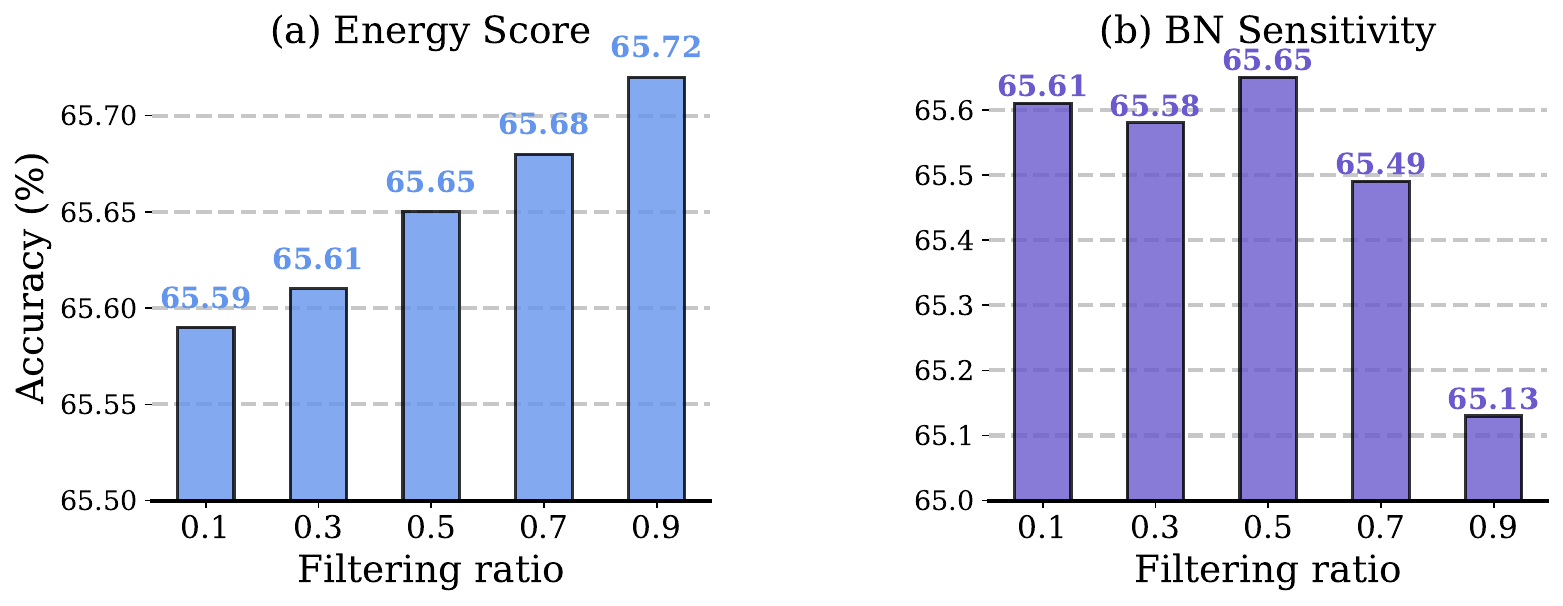}
    \caption{Accuracy vs filtering strength with data-free PTQ. }
    \label{fig_ablation}
    \vspace{-1.5em}
\end{figure}

\fakeparagraph{Energy Score Filtering Strength. }
Given a fixed number of synthetic data ($N$) used for quantization and a percentage $r$ representing how many images are filtered out, the total amount of generated images is $\frac{N}{1-r}$. We test different options of $r$ from $\{0.1, 0.3, 0.5, 0.7, 0.9\}$ to test the effect of energy score filtering. We choose ResNet-18 and use W2A4 data-free PTQ (with Genie-M) for evaluation. The test performance is shown in \autoref{fig_ablation}(a). We generally find that a higher filtering ratio leads to better test accuracy. However, it will also increase the number of generated images. Nevertheless, the accuracy variation is rather small.

\fakeparagraph{BN Sensitivity Filtering Strength. }
We run the same test for the BN sensitivity filtering mechanism. The results are shown in \autoref{fig_ablation}(b). Unlike the energy score, setting the ratio to 50\% has the best performance. We hypothesize that the networks need more diverse synthetic data to effectively quantize a model.

\section{Conclusion}

In this paper, we have introduced \genq, a novel attempt to synthesize images with text-to-image models for data-scarce quantization. Our \genq generates images in both data free (no real data) and data scarce (few real data) regimes and refines the images with several filtering mechanisms and a token embedding learning algorithm. Extensive experiments show that \genq establishes a new state of the art in both PTQ and QAT.  

\section*{Acknowledgment}
This work was supported in part by CoCoSys, a JUMP2.0 center sponsored by DARPA and SRC, the National Science Foundation (CAREER Award, Grant \#2312366, Grant \#2318152), and the DoE MMICC center SEA-CROGS (Award \#DE-SC0023198)


%
%
\bibliographystyle{splncs04}
\bibliography{main}

\clearpage
\appendix

\section{Technical Details}

\fakeparagraph{Textual Inversion in Stable Diffusion. }
In \cref{sec_dsq}, we optimize a token embedding for data-scarce quantization. Specifically, this technique is based on the Textual Inversion~\cite{gal2022image} where a token embedding is optimized for accurate recreation of an object from 3-5 images. 
To optimize the token embedding, we freeze all other parts of the Stable Diffusion including all the parameters from the encoder, decoder, text encoder, and the denoising U-Net. The training objective is the same as the training of Stable Diffusion, \ie diffuse the real images latent into noisy latent and learn how to denoise it except we only update the text embedding (\cf \cref{eq:fewshot_opt}). 
Note that the original textual inversion is designed to find the token of the exact \textit{object} while we have multiple images from multiple classes. Hence, our objective is to find some token embedding that can universally describe the traits of the ImageNet data, aiming to learn distinct distribution characteristics. 

\fakeparagraph{Implementation of Token Embedding Learning. }
To optimize $\mathtt{\{S\}}$, we optimize it for 50k iterations with a batch size of 32 and use Adam optimizer with a constant learning rate of $5e-4$, following the convention of implementation~\cite{gal2022image}.
Note that for each batch, we construct 32 different prompts since each image may have a unique label name. 

\section{Prompt Template}

We use the following template to generate the prompt:

\begin{enumerate}
\item \texttt{  photo of a \{C\}.}
\item \texttt{  rendering of a \{C\}.}
\item \texttt{  cropped photo of the \{C\}.}
\item \texttt{  the photo of a \{C\}.}
\item \texttt{  photo of a clean \{C\}.}
\item \texttt{  photo of a dirty \{C\}.}
\item \texttt{  dark photo of the \{C\}.}
\item \texttt{  photo of my \{C\}.}
\item \texttt{  photo of the cool \{C\}.}
\item \texttt{  close-up photo of a \{C\}.}
\item \texttt{  bright photo of the \{C\}.}
\item \texttt{  cropped photo of a \{C\}.}
\item \texttt{  photo of the \{C\}.}
\item \texttt{  good photo of the \{C\}.}
\item \texttt{  photo of one \{C\}.}
\item \texttt{  close-up photo of the \{C\}.}
\item \texttt{  rendition of the \{C\}.}
\item \texttt{  photo of the clean \{C\}.}
\item \texttt{  rendition of a \{C\}.}
\item \texttt{  photo of a nice \{C\}.}
\item \texttt{  good photo of a \{C\}.}
\item \texttt{  photo of the nice \{C\}.}
\item \texttt{  photo of the small \{C\}.}
\item \texttt{  photo of the weird \{C\}.}
\item \texttt{  photo of the large \{C\}.}
\item \texttt{  photo of a cool \{C\}.}
\item \texttt{  photo of a small \{C\}.}
\end{enumerate}

\end{document}